\documentclass[runningheads]{llncs}
\usepackage{float}
\usepackage{bm}
\usepackage{mathtools}  
\usepackage{diffcoeff}[=v4]
\usepackage{amssymb}
\usepackage{url}
\usepackage[inline]{enumitem}
\usepackage{varwidth}
\usepackage{capt-of}
\usepackage{wrapfig,lipsum,booktabs}
\usepackage{comment}
\usepackage{xspace}
\usepackage{amsmath}
\usepackage{amssymb}
\usepackage{booktabs}
\usepackage{tabularx}
\usepackage{graphicx}
\usepackage{cite}
\usepackage{multirow}
\usepackage{subfigure}
\usepackage{diagbox}
\usepackage{paralist}
\usepackage{xcolor}
\usepackage[normalem]{ulem}
\usepackage{hyperref}
\hypersetup{
  colorlinks,
  citecolor=violet,
  linkcolor=red,
  urlcolor=blue
}

\def\eg{\emph{e.g.}}

\def\ie{\emph{i.e.}}
\def\vs{\emph{v.s.}}

\usepackage{bbding}
\newcommand\blfootnote[1]{%
  \begingroup
  \renewcommand\thefootnote{}\footnote{#1}%
  \addtocounter{footnote}{-1}%
  \endgroup
}

\begin{document}
\title{EndoSurf: Neural Surface Reconstruction of Deformable Tissues with Stereo Endoscope Videos}
\titlerunning{EndoSurf 3D Reconstruction}

\author{\textsuperscript{*}Ruyi Zha\inst{1} \Envelope \and
\textsuperscript{*}Xuelian Cheng\inst{2,4,5} \and
Hongdong Li\inst{1} \and
Mehrtash Harandi\inst{2} \and
Zongyuan Ge \inst{2,3,4,5,6}
}
\index{Ruyi, Zha}
\index{Xuelian, Cheng} 
\index{Hongdong, Li}
\index{Mehrtash, Harandi} 
\index{Zongyuan, Ge}

\authorrunning{R. Zha et al.}
%
\institute
{Australian National University, Canberra, Australia, 
\and Faculty of Engineering, Monash University, Melbourne, Australia
\and Faculty of IT, Monash University, Melbourne, Australia 
\and AIM for Health Lab, Monash University, Melbourne, Australia 
\and Monash Medical AI, Monash University, Melbourne, Australia 
\and Airdoc-Monash Research Lab, Monash University, Melbourne, Australia 
}

\maketitle              
\begin{abstract}
 Reconstructing soft tissues from stereo endoscope videos is an essential prerequisite for many medical applications. Previous methods struggle to produce high-quality geometry and appearance due to their inadequate representations of 3D scenes. To address this issue, we propose a novel neural-field-based method, called \textit{EndoSurf}, which effectively learns to represent a deforming surface from an RGBD sequence. In EndoSurf, we model surface dynamics, shape, and texture with three neural fields. First, 3D points are transformed from the observed space to the canonical space using the deformation field. The signed distance function (SDF) field and radiance field then predict their SDFs and colors, respectively, with which RGBD images can be synthesized via differentiable volume rendering. We constrain the learned shape by tailoring multiple regularization strategies and disentangling geometry and appearance. Experiments on public endoscope datasets demonstrate that EndoSurf significantly outperforms existing solutions, particularly in reconstructing high-fidelity shapes. Code is available at \url{https://github.com/Ruyi-Zha/endosurf.git}.

\end{abstract}
\blfootnote{* Equal contribution.}
\section{Introduction}
\label{sec:introduction}

Surgical scene reconstruction using stereo endoscopes is crucial to Robotic-Assisted Minimally Invasive Surgery (RAMIS). It aims to recover a 3D model of the observed tissues from a stereo endoscope video. Compared with traditional 2D monitoring, 3D reconstruction offers notable advantages because it allows users to observe the surgical site from any viewpoint. Therefore, it dramatically benefits downstream medical applications such as surgical navigation~\cite{overley2017navigation}, surgeon-centered augmented reality~\cite{nicolau2011augmented}, and virtual reality~\cite{chong2021virtual}. General reconstruction pipelines first estimate depth maps with stereo-matching~\cite{li2021revisiting, cheng2020hierarchical,cheng2022deep} and then fuse RGBD images into a 3D model~\cite{li2020super,long2021dssr,song2017dynamic,zhou2021emdq,newcombe2015dynamicfusion}. Our work focuses on the latter, \ie, how to accurately reconstruct the shape and appearance of deforming surfaces from RGBD sequences.

Existing approaches represent a 3D scene in two ways: discretely or continuously. Discrete representations include point clouds~\cite{li2020super,long2021dssr,song2017dynamic,zhou2021emdq} and mesh grids~\cite{newcombe2015dynamicfusion}. Additional warp fields~\cite{gao2019surfelwarp} are usually utilized to compensate for tissue deformation. Discrete representation methods produce surfaces efficiently due to their sparsity property. However, this property also limits their ability to handle complex high-dimensional changes, \eg, non-topology deformation and color alteration resulting from cutting or pulling tissues.

\begin{figure}[t]
    \centering
    \includegraphics[width=\linewidth]{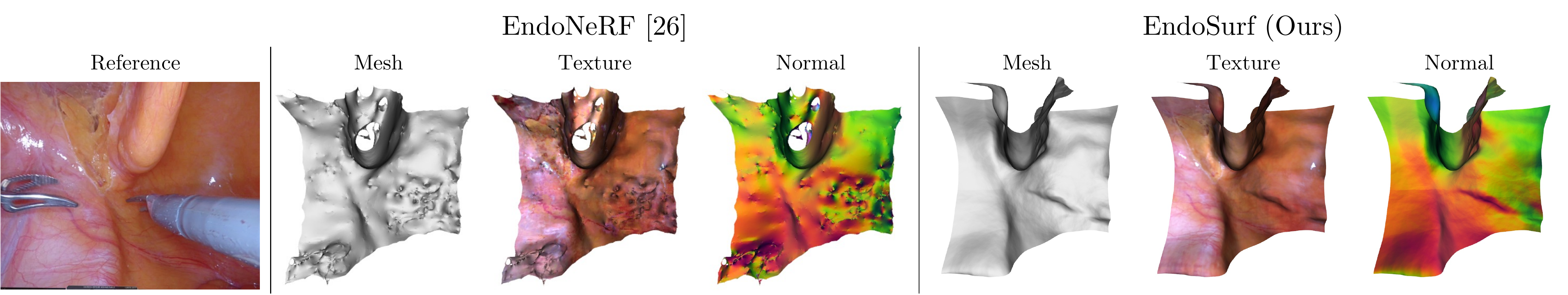}
    \caption{3D meshes extracted from EndoNeRF~\cite{wang2022neural} and our method. EndoNeRF cannot recover a smooth and accurate surface even with post-processing filters.}
    \label{fig:simple_demo}
\end{figure}

Recently, continuous representations have become popular with the blossoming of neural fields, \ie, neural networks that take space-time inputs and return the required quantities. Neural-field-based methods~\cite{wang2022neural,yariv2020multiview,pumarola2021d,wang2021neus,niemeyer2020differentiable,oechsle2021unisurf} exploit deep neural networks to implicitly model complex geometry and appearance, outperforming discrete-representation-based methods. A good representative is EndoNeRF~\cite{wang2022neural}. It trains two neural fields: one for tissue deformation and the other for canonical density and color. EndoNeRF can synthesize reasonable RGBD images with post-processing filters. However, the ill-constrained properties of the density field deter the network from learning a solid surface shape. Fig.~\ref{fig:simple_demo} shows that EndoNeRF can not accurately recover the surface even with filters. While there have been attempts to parameterize other geometry fields, \eg, occupancy fields~\cite{niemeyer2020differentiable,oechsle2021unisurf} and signed distance function (SDF) fields~\cite{yariv2020multiview,wang2021neus}, they hypothesize static scenes and diverse viewpoints. Adapting them to surgical scenarios where surfaces undergo deformation and camera movement is confined is non-trivial.

We propose EndoSurf: neural implicit fields for \textbf{Endo}scope-based \textbf{Surf}ace reconstruction, a novel neural-field-based method that effectively learns to represent dynamic scenes. Specifically, we model deformation, geometry, and appearance with three separate multi-layer perceptrons (MLP). The deformation network transforms points from the observation space to the canonical space. The geometry network represents the canonical scene as an SDF field. Compared with density, SDF is more self-contained as it explicitly defines the surface as the zero-level set. We enforce the geometry network to learn a solid surface by designing various regularization strategies. Regarding the appearance network, we involve positions and normals as extra clues to disentangle the appearance from the geometry. Following~\cite{wang2021neus}, we adopt unbiased volume rendering to synthesize color images and depth maps. The network is optimized with gradient descent by minimizing the error between the real and rendered results. We evaluate EndoSurf quantitatively and qualitatively on public endoscope datasets. Our work demonstrates superior performance over existing solutions, especially in reconstructing smooth and accurate shapes. 

\section{Method}

\subsection{Overview}

\subsubsection{Problem setting}

Given a stereo video of deforming tissues, we aim to reconstruct the surface shape $\mathcal{S}$ and texture $\mathcal{C}$. Similar to EndoNeRF~\cite{wang2022neural}, we take as inputs a sequence of frame data $\{(\mathbf{I}_{i}, \mathbf{D}_{i}, \mathbf{M}_{i}, \mathbf{P}_{i}, t_{i})\}_{i=1}^{T}$. Here $T$ stands for the total number of frames. $\mathbf{I}_{i}\in\mathbb{R}^{H\times W\times3}$ and $\mathbf{D}_i\in \mathbb{R}^{H\times W}$ refer to the $i$-th left RGB image and depth map with height $H$ and width $W$. Foreground mask $\mathbf{M}_{i}\in\mathbb{R}^{H\times W}$ is utilized to exclude unwanted pixels, such as surgical tools, blood, and smoke. Projection matrix $\mathbf{P}_{i}\in{\mathbb{R}}^{4\times4}$ maps 3D coordinates to 2D pixels. $t_{i}=i/T$ is each frame's timestamp normalized to $[0,1]$. While stereo matching, surgical tool tracking, and pose estimation are also practical clinical concerns, in this work we prioritize 3D reconstruction and thus take depth maps, foreground masks, and projection matrices as provided by software or hardware solutions.

\subsubsection{Pipeline}
\begin{figure}[t]
    \centering
    \includegraphics[width=1.0\linewidth]{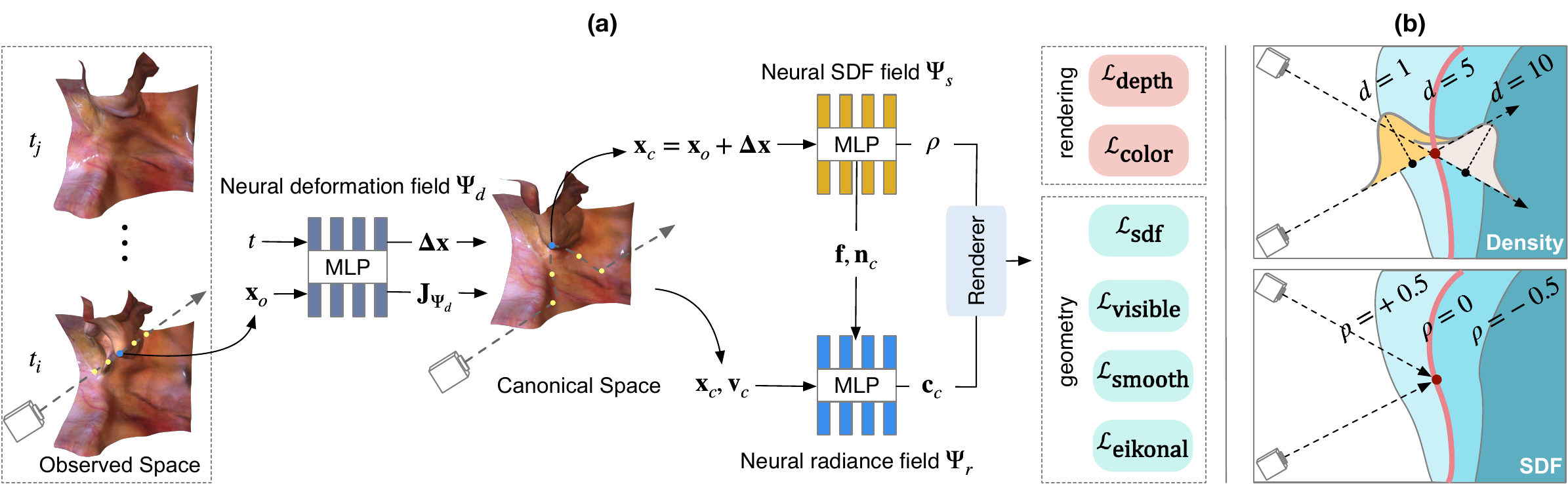} 
    \caption{(a) The overall pipeline of EndoSurf. (b) Density field \vs~SDF field.  Red lines represent surfaces. The density field is depth ambiguous, while the SDF field clearly defines the surface as the zero-level set.}
    \label{fig:architecture}
\end{figure}

Fig.~\ref{fig:architecture} (a) illustrates the overall pipeline of our approach. Similar to~\cite{wang2022neural,wang2021neus}, we incorporate our EndoSurf network into a volume rendering scheme. Specifically, we begin by adopting a mask-guided sampling strategy~\cite{wang2022neural} to select valuable pixels from a video frame. We then cast 3D rays from these pixels and hierarchically sample points along the rays~\cite{wang2021neus}. The EndoSurf network utilizes these sampled points and predicts their SDFs and colors. After that, we adopt the unbiased volume rendering method~\cite{wang2021neus} to synthesize pixel colors and depths used for network training. We tailor loss functions to enhance the network's learning of geometry and appearance. In the following subsections, we will describe the EndoSurf network (cf. Sec.~\ref{sec:endosurf}) and the optimization process (cf. Sec~\ref{sec:optimization}) in detail.

\subsection{EndoSurf: Representing scenes as deformable neural fields}
\label{sec:endosurf}

We represent a dynamic scene as canonical neural fields warped to an observed pose. Separating the learning of deformation and canonical shapes has been proven more effective than directly modeling dynamic shapes~\cite{pumarola2021d}. Particularly, we propose a neural deformation field $\mathbf{\Psi}_{d}$ to transform 3D points from the observed space to the canonical space. The geometry and appearance of the canonical scene are described by a neural SDF field $\mathbf{\Psi}_{s}$ and a neural radiance field $\mathbf{\Psi}_{r}$, respectively. All neural fields are modeled with MLPs with position encoding~\cite{mildenhall2020nerf,tancik2020fourier}.

\subsubsection{Neural deformation field}

Provided a 3D point $\mathbf{x}_{o}\in\mathbb{R}^{3}$ in the observed space at time $t\in[0,1]$, the neural deformation field $\mathbf{\Psi}_{d}(\mathbf{x}_{o},t)\mapsto \mathbf{\Delta x}$ returns the displacement $\mathbf{\Delta x}\in\mathbb{R}^{3}$ that transforms $\mathbf{x}_{o}$ to its canonical position $\mathbf{x}_{c}=\mathbf{x}_{o}+\mathbf{\Delta x}$. The canonical view direction $\mathbf{v}_{c}\in \mathbb{R}^{3}$ of point $\mathbf{x}_{c}$ can be obtained by transforming the raw view direction $\mathbf{v}_{o}$ with the Jacobian of the deformation field $\mathbf{J}_{\mathbf{\Psi}_{d}}(\mathbf{x}_{o})=\diffp{\mathbf{\Psi}_{d}}/{\mathbf{x}_{o}}$, \ie, $\mathbf{v}_{c} = (\mathbf{I} + \mathbf{J}_{\mathbf{\Psi}_{d}}(\mathbf{x}_{o})) \mathbf{v}_{o}$.

\subsubsection{Neural SDF field}

The shape of the canonical scene is represented by a neural field $\mathbf{\Psi}_{s}(\mathbf{x}_{c}) \mapsto (\rho, \mathbf{f})$ that maps a spatial position $\mathbf{x}_{c}\in \mathbb{R}^{3}$ to its signed distance function $\rho\in \mathbb{R}$ and a geometry feature vector $\mathbf{f}\in \mathbb{R}^{F}$ with feature size $F$.

In 3D vision, SDF is the orthogonal distance of a point $\mathbf{x}$ to a watertight object's surface, with the sign determined by whether or not $\mathbf{x}$ is outside the object. In our case, we slightly abuse the term SDF since we are interested in a segment of an object rather than the whole thing. We extend the definition of SDF by imagining that the surface of interest divides the surrounding space into two distinct regions, as shown in Fig.~\ref{fig:architecture} (b). SDF is positive if $\mathbf{x}$ falls into the region which includes the camera and negative if it is in the other. As $\mathbf{x}$ approaches the surface, the SDF value gets smaller until it reaches zero at the surface. Therefore, the surface of interest $\mathcal{S}$ is the zero-level set of SDF, \ie, $\mathcal{S}=\{\mathbf{p}\in\mathbb{R}^3|\mathbf{\Psi}_{s}(\mathbf{p})=0\}$. 

Compared with the density field used in~\cite{wang2022neural}, the SDF field provides a more precise representation of surface geometry because it explicitly defines the surface as the zero-level set. The density field, however, encodes the probability of an object occupying a position, making it unclear which iso-surface defines the object's boundary. As a result, density-field-based methods~\cite{wang2022neural,mildenhall2020nerf} can not directly identify a depth via ray marching but rather render it by integrating the depths of sampled points with density-related weights. Such a rendering method can lead to potential depth ambiguity, \ie, camera rays pointing to the same surface produce different surface positions (Fig.~\ref{fig:architecture} (b)). 

Given a surface point $\mathbf{p}_{c}\in\mathbb{R}^3$ in the canonical space, the surface normal $\mathbf{n}_{c}\in\mathbb{R}^3$ is the gradient of the neural SDF field $\mathbf{\Psi}_{s}$:  $\mathbf{n}_{c}=\nabla_{\mathbf{\Psi}_{s}}(\mathbf{p}_{c})$. Normal $\mathbf{n}_{o}$ of the deformed surface point $\mathbf{p}_{o}$ can also be obtained with the chain rule.

\subsubsection{Neural radiance field}

We model the appearance of the canonical scene as a neural radiance field $\mathbf{\Psi}_{r}(\mathbf{x}_{c}, \mathbf{v}_{c}, \mathbf{n}_{c}, \mathbf{f})\mapsto \mathbf{c}_{c}$ that returns the color $ \mathbf{c}_{c}\in \mathbb{R}^3$ of a viewpoint $(\mathbf{x}_{c}, \mathbf{v}_{c})$. Unlike~\cite{wang2022neural,mildenhall2020nerf}, which only take the view direction $\mathbf{v}_{c}$ and feature vector $\mathbf{f}$ as inputs, we also feed the normal $\mathbf{n}_{c}$ and position $\mathbf{x}_{c}$ to the radiance field as extra geometric clues. Although the feature vector implies the normal and position information, it is validated that directly incorporating them benefits the disentanglement of geometry, \ie, allowing the network to learn appearance independently from the geometry~\cite{yariv2020multiview,wang2021neus}.

\subsection{Optimization}
\label{sec:optimization}

\subsubsection{Unbiased volume rendering}

Given a camera ray $\mathbf{r}(h)=\mathbf{o}_{o}+h\mathbf{v}_{o}$ at time $t$ in the observed space, we sample $N$ points $\mathbf{x}_{i}$ in a hierarchical manner along this ray~\cite{wang2021neus} and predict their SDFs $\rho_{i}$ and colors $\mathbf{c}_{i}$ via EndoSurf. The color $\hat{\mathbf{C}}$ and depth $\hat{\mathbf{D}}$ of the ray can be approximated by unbiased volume rendering~\cite{wang2021neus}:

\begin{equation}
\hat{\mathbf{C}}(\mathbf{r}(h))=\sum\nolimits_{i=1}^{N} T_{i} \alpha_{i} \mathbf{c}_{i},\quad \hat{\mathbf{D}}(\mathbf{r}(h))=\sum\nolimits_{i=1}^{N} T_{i} \alpha_{i} h_{i},
\end{equation}
where $T_{i}=\prod\nolimits_{j=1}^{i-1}(1-\alpha_{j})$, $\alpha_{i}=\max((\phi(\rho_{i})-\phi(\rho_{i+1}))/\phi({\rho_{i}}), 0)$ and $\phi(\rho)=(1+e^{-\rho/s})^{-1}$. Note that $s$ is a trainable standard deviation, which approaches zero as the network training converges.

\subsubsection{Loss}
We train the network with two objectives: 1) to minimize the difference between the actual and rendered results and 2) to impose constraints on the neural SDF field such that it aligns with its definition. Accordingly, we design two categories of losses: rendering constraints and geometry constraints:

\begin{equation}
    \mathcal{L} = {\underbrace {\Big(\lambda_1\mathcal{L}_{\textrm{color}} + \lambda_2\mathcal{L}_{\textrm{depth}}\Big)}_\text{rendering}} + {\underbrace {\Big( \lambda_3\mathcal{L}_{\textrm{eikonal}}+\lambda_4\mathcal{L}_{\textrm{sdf}} + \lambda_5\mathcal{L}_{\textrm{visible}}+ \lambda_6\mathcal{L}_{\textrm{smooth}} \Big)}_\text{geometry}},
\label{eq:total_loss}
\end{equation}
where $\lambda_{i=1,\cdots,6}$ are balancing weights. The rendering constraints include the color reconstruction loss $\mathcal{L}_{\textrm{color}}$ and depth reconstruction loss $\mathcal{L}_{\textrm{depth}}$:

\begin{equation}
    \mathcal{L}_{\textrm{color}}=\sum\limits_{\mathbf{r}\in\mathcal{R}} \Vert M(\mathbf{r})(\hat{\mathbf{C}}(\mathbf{r})-\mathbf{C}(\mathbf{r})) \Vert_{1}, \mathcal{L}_{\textrm{depth}}=\sum\limits_{\mathbf{r}\in\mathcal{R}} \Vert M(\mathbf{r})(\hat{\mathbf{D}}(\mathbf{r})-\mathbf{D}(\mathbf{r})) \Vert_{1},
\end{equation}
where $M(\mathbf{r})$, $\{\hat{\mathbf{C}}, \hat{\mathbf{D}}\}$, $\{\mathbf{C}, \mathbf{D}\}$ and $\mathcal{R}$ are ray masks, rendered colors and depths, real colors and depths, and ray batch, respectively.

We regularize the neural SDF field $\mathbf{\Psi}_{s}$ with four losses: Eikonal loss $\mathcal{L}_{\textrm{eikonal}}$, SDF loss $\mathcal{L}_{\textrm{sdf}}$, visibility loss $\mathcal{L}_{\textrm{visible}}$, and smoothness loss $\mathcal{L}_{\textrm{smooth}}$.

\begin{equation}
\begin{gathered}
    \mathcal{L}_{\textrm{eikonal}}=\sum\limits_{\mathbf{x}\in\mathcal{X}}(\Vert \nabla_{\mathbf{\Psi}_{s}}(\mathbf{x}) \Vert_{2}-1)^{2}, \mathcal{L}_{\textrm{sdf}}=\sum\limits_{\mathbf{p}\in\mathcal{D}} \Vert \mathbf{\Psi}_{s}(\mathbf{p}) \Vert_{1}, \\
    \mathcal{L}_{\textrm{visible}}=\sum\limits_{\mathbf{p}\in\mathcal{D}}\max(\langle \nabla_{\mathbf{\Psi}_{s}}(\mathbf{p}), \mathbf{v}_{c} \rangle,0), \mathcal{L}_{\textrm{smooth}}=\sum\limits_{\mathbf{p}\in\mathcal{D}}\Vert \nabla_{\mathbf{\Psi}_{s}}(\mathbf{p})-\nabla_{\mathbf{\Psi}_{s}}(\mathbf{p}+\bm{\epsilon}) \Vert_{1}.
\end{gathered}
\end{equation}
Here the Eikonal loss $\mathcal{L}_{\textrm{eikonal}}$~\cite{gropp2020implicit} encourages $\mathbf{\Psi}_{s}$ to satisfy the Eikonal equation~\cite{crandall1983viscosity}. Points $\mathbf{x}$ are sampled from the canonical space $\mathcal{X}$. The SDF loss  $\mathcal{L}_{\textrm{sdf}}$ restricts the SDF value of points lying on the ground truth depths $\mathcal{D}$ to zero. The visibility loss $\mathcal{L}_{\textrm{visible}}$ limits the angle between the canonical surface normal and the viewing direction $\mathbf{v}_{c}$ to be greater than 90 degrees. The smoothness loss $\mathcal{L}_{\textrm{smooth}}$ encourages a surface point and its neighbor to be similar, where $\bm{\epsilon}$ is a random uniform perturbation.

\section{Experiments}

\subsection{Experiment settings}

\subsubsection{Datasets and evaluation}

We conduct experiments on two public endoscope datasets, namely ENDONERF~\cite{wang2022neural} and SCARED~\cite{allan2021stereo} 
(See statistical details in the supplementary material).
ENDONERF provides two cases of in-vivo prostatectomy data with estimated depth maps~\cite{li2021revisiting} and manually labeled tool masks. SCARED~\cite{allan2021stereo} collects the ground truth RGBD images of five porcine cadaver abdominal anatomies. We pre-process the datasets by normalizing the scene into a unit sphere and splitting the frame data into 7:1 training and test sets.

Our approach is compared with EndoNeRF~\cite{wang2022neural}, the state-of-the-art neural-field-based method. There are three outputs for test frames: RGB images, depth maps, and 3D meshes. The first two outputs are rendered the same way as the training process. We use marching cubes~\cite{lorensen1987marching} to extract 3D meshes from the density and SDF fields. The threshold is set to 5 for the density field and 0 for the SDF field. See the supplementary material for the validation of threshold selection. Five evaluation metrics are used: PSNR, SSIM, LPIPS, RMSE, and point cloud distance (PCD). The first three metrics assess the similarity between the actual and rendered RGB images\cite{wang2022neural}, while RMSE and PCD measure depth map~\cite{cheng2020hierarchical,cheng2022deep,long2021dssr} and 3D mesh~\cite{cai2022neural,bozic2020deepdeform} reconstruction quality, respectively.

\subsubsection{Implementation details}

We train neural networks per scene, \ie, one model for each case. All neural fields consist of 8-layer 256-channel MLPs with a skip connection at the 4th layer. Position encoding frequencies in all fields are 6, except those in the radiance field are 10 and 4 for location and direction, respectively. The SDF network is initialized~\cite{atzmon2020sal} for better training convergence. We use Adam optimizer~\cite{kingma2014adam} with a learning rate of 0.0005, which warms up for 5k iterations and then decays with a rate of 0.05. We sample 1024 rays per batch and 64 points per ray. The initial standard deviation $s$ is 0.3. The weights in Eq.~\ref{eq:total_loss} are $\lambda_{1}=1.0$, $\lambda_{2}=1.0$, $\lambda_{3}=0.1$, $\lambda_{4}=1.0$, $\lambda_{5}=0.1$ and $\lambda_{6}=0.1$. We train our model with $100K$ iterations for 9 hours on an NVIDIA RTX 3090 GPU.

\subsection{Qualitative and quantitative results}

\begin{table}[t]
\tiny
\centering
\caption{Quantitative metrics of appearance (PSNR/SSIM/LPIPS) and geometry (RMSE/PCD) on two datasets. The unit for RMSE/PCD is millimeter.}
\label{tab:results}
\resizebox{\columnwidth}{!}{%
\begin{tabular}{@{}c|ccccc|ccccc@{}}
\toprule
Methods & \multicolumn{5}{c|}{EndoNeRF~\cite{wang2022neural}} & \multicolumn{5}{c}{EndoSurf (Ours)} \\ \midrule
Metrics & PSNR$\uparrow$ & SSIM$\uparrow$ & {LPIPS$\downarrow$} & RMSE$\downarrow$ & PCD$\downarrow$ & PSNR$\uparrow$ & SSIM$\uparrow$ & {LPIPS$\downarrow$} & RMSE$\downarrow$ & PCD$\downarrow$ \\ \midrule
ENDONERF-cutting & 34.186 & 0.932 & 0.151 & 0.930 & 1.030 & \textbf{34.981} & \textbf{0.953} & \textbf{0.106} & \textbf{0.835} & \textbf{0.559} \\
ENDONERF-pulling & 34.212 & 0.938 & 0.161 & 1.485 & 2.260 & \textbf{35.004} & \textbf{0.956} & \textbf{0.120} & \textbf{1.165} & \textbf{0.841} \\
SCARED-d1k1 & 24.365 & 0.763 & 0.326 & 0.697 & 2.982 & \textbf{24.395} & \textbf{0.769} & \textbf{0.319} & \textbf{0.522} & \textbf{0.741} \\
SCARED-d2k1 & 25.733 & 0.828 & \textbf{0.240} & 0.583 & 1.788 & \textbf{26.237} & \textbf{0.829} & 0.254 & \textbf{0.352} & \textbf{0.515} \\
SCARED-d3k1 & 19.004 & 0.599 & 0.467 & 1.809 & 3.244 & \textbf{20.041} & \textbf{0.649} & \textbf{0.441} & \textbf{1.576} & \textbf{1.091} \\
SCARED-d6k1 & 24.041 & 0.833 & 0.464 & 1.194 & 3.268 & \textbf{24.094} & \textbf{0.866} & \textbf{0.461} & \textbf{1.065} & \textbf{1.331} \\
SCARED-d7k1 & 22.637 & 0.813 & 0.312 & 2.272 & 3.465 & \textbf{23.421} & \textbf{0.861} & \textbf{0.282} & \textbf{2.123} & \textbf{1.589} \\ \midrule
Average & 26.311 & 0.815 & 0.303 & 1.281 & 2.577 & \textbf{26.882} & \textbf{0.840} & \textbf{0.283} & \textbf{1.091} & \textbf{0.952} \\ \bottomrule
\end{tabular}%
}
\end{table}

\begin{figure}[t]
    \centering
    \includegraphics[width=\linewidth]{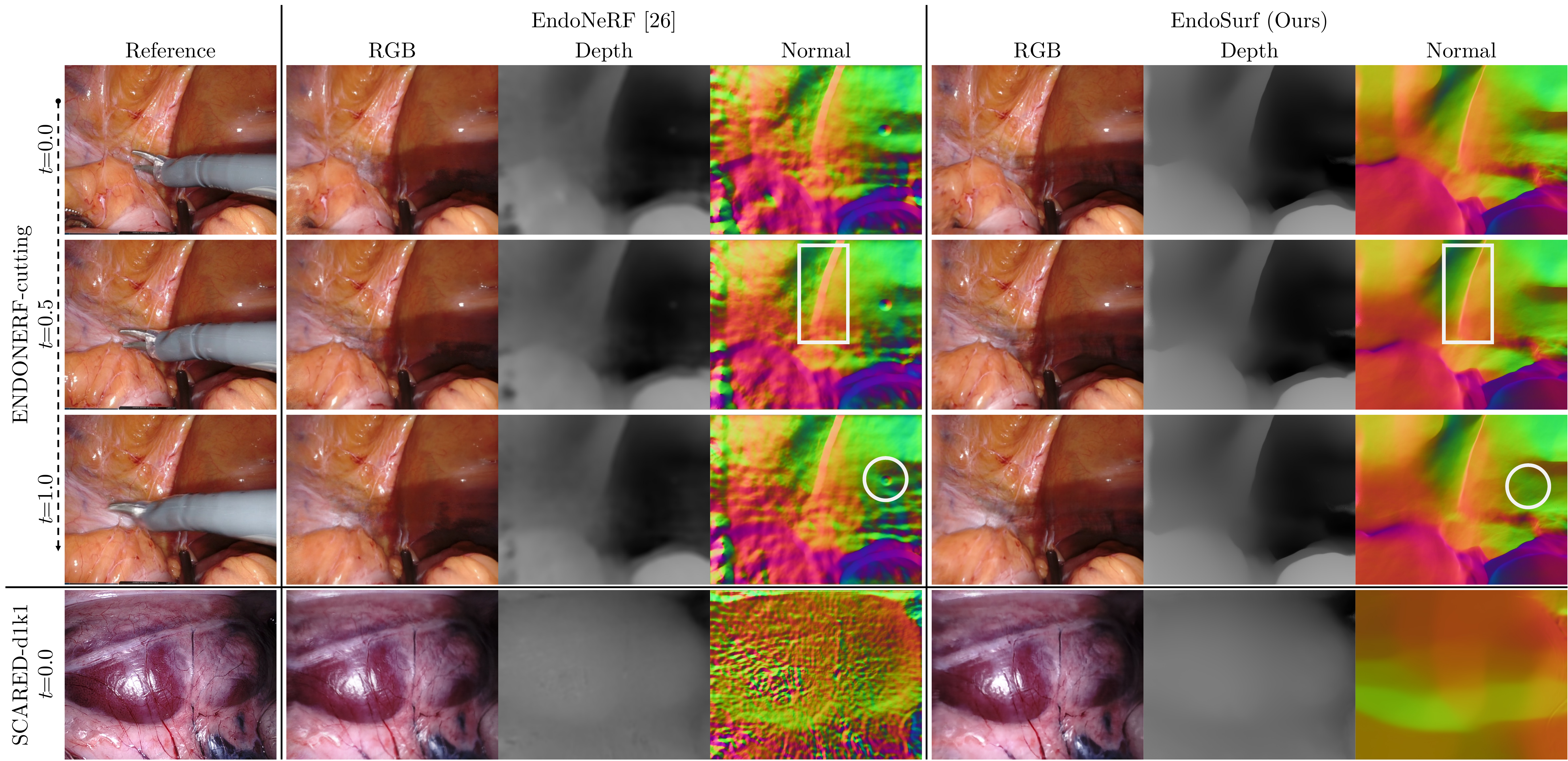}
    \caption{2D rendering results on the dynamic case ``ENDONERF-cutting" and static case ``SCARED-d1k1". Our method yields high-quality depth and normal maps, whereas those of EndoNeRF exhibit jagged noise, over-smoothed edges (white boxes), and noticeable artifacts (white rings).}
    \label{fig:render_results}
\end{figure}

As listed in Table~\ref{tab:results}, EndoSurf yields superior results against EndoNeRF. On the one hand, EndoSurf produces better appearance quality than EndoNeRF by $\uparrow$ 0.571 PSNR, $\uparrow$ 0.025 SSIM, and $\uparrow$ 0.020 LPIPS. On the other hand, EndoSurf dramatically outperforms EndoNeRF in terms of geometry recovery by $\downarrow$ 0.190 RMSE and $\downarrow$ 1.625 PCD. Note that both methods perform better on ENDONERF than on SCARED. This is because ENDONERF fixes the camera pose, leading to easier network fitting.

Fig.~\ref{fig:render_results} shows the 2D rendering results. While both methods synthesize high-fidelity RGB images, only EndoSurf succeeds in recovering depth maps with a smoother shape, more details, and fewer artifacts. First, the geometry constraints in EndoSurf prevent the network from overfitting depth supervision, suppressing rough surfaces as observed in the EndoNeRF's normal maps. Second, the brutal post-processing filtering in EndoNeRF cannot preserve sharp details (white boxes in Fig.~\ref{fig:render_results}). Moreover, the texture and shape of EndoNeRF are not disentangled, causing depth artifacts in some color change areas (white rings in Fig.~\ref{fig:render_results}).

\begin{figure}[t]
    \centering
    \includegraphics[width=\linewidth]{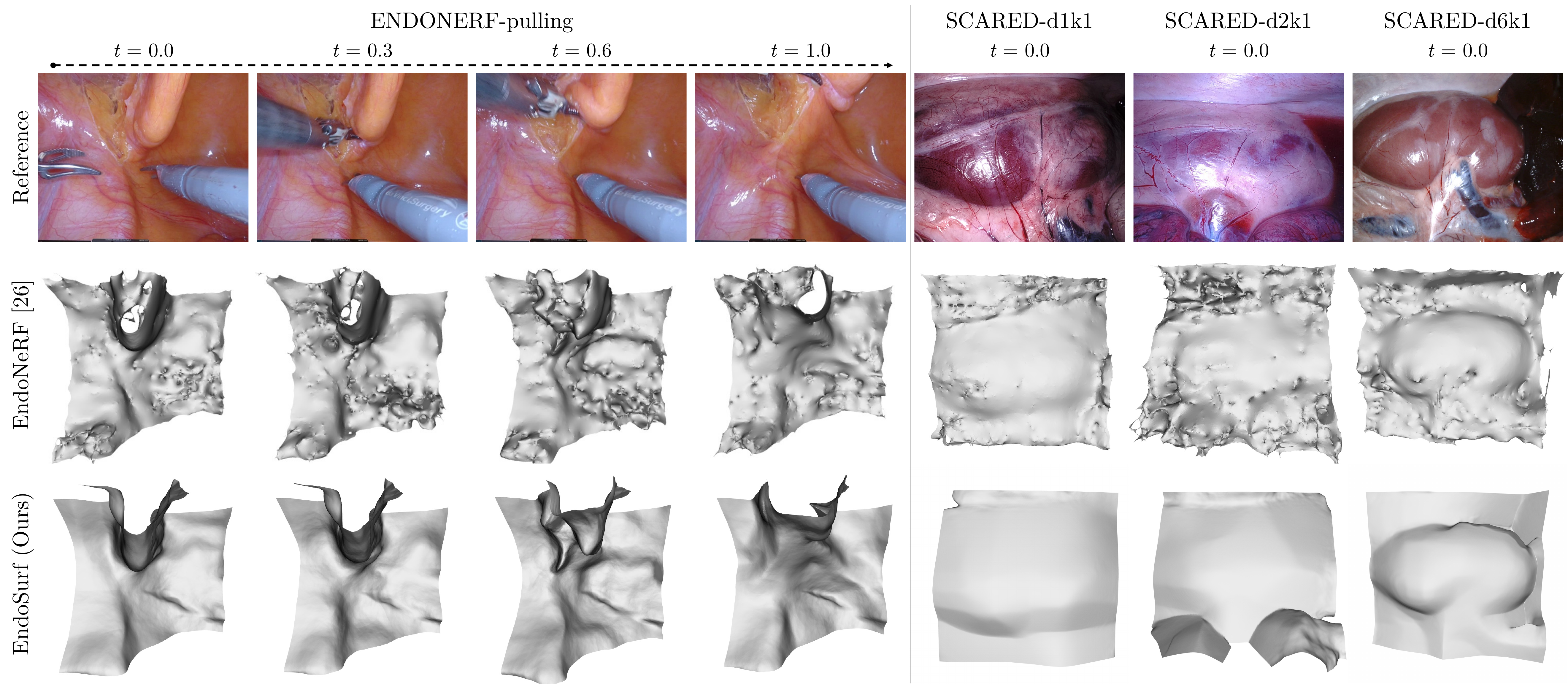}
    \caption{Extracted meshes on one dynamic scene and three static scenes. Our method produces accurate and smooth surfaces.}
    \label{fig:marching_cubes}
\end{figure}

Fig.~\ref{fig:marching_cubes} depicts the shapes of extracted 3D meshes. Surfaces reconstructed by EndoSurf are accurate and smooth, while those from EndoNeRF are quite noisy. There are two reasons for the poor quality of EndoNeRF's meshes. First, the density field without regularization tends to describe the scene as a volumetric fog rather than a solid surface. Second, the traditional volume rendering causes discernible depth bias~\cite{wang2021neus}. In contrast, we force the neural SDF field to conform to its definition via multiple geometry constraints. Furthermore, we use unbiased volume rendering to prevent depth ambiguity~\cite{wang2021neus}.

\begin{figure}
    \centering
    \includegraphics[width=\linewidth]{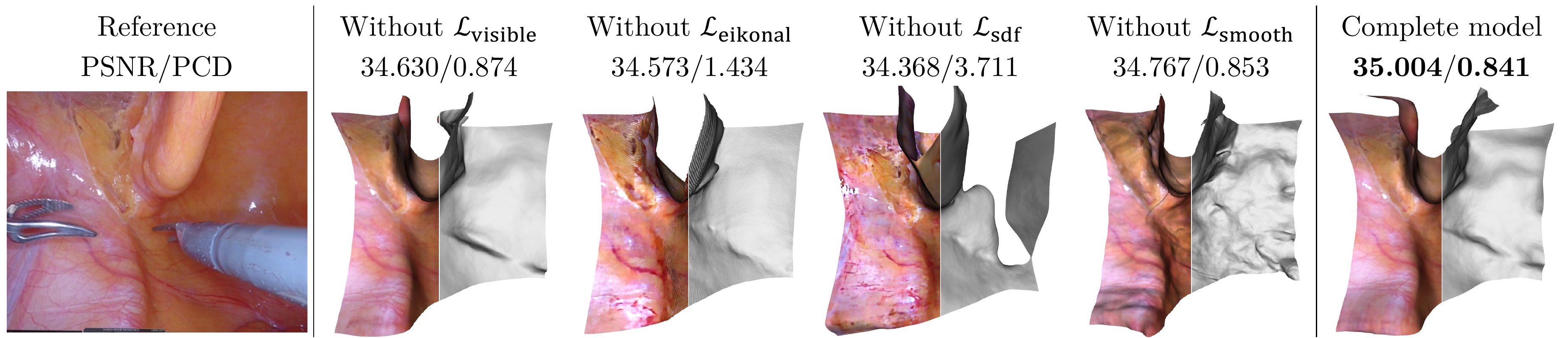}
    \caption{Ablation study on four geometry constraints, \ie, visibility loss $\mathcal{L}_{\textrm{visible}}$, Eikonal loss $\mathcal{L}_{\textrm{eikonal}}$, SDF loss $\mathcal{L}_{\textrm{sdf}}$, and smoothness loss $\mathcal{L}_{\textrm{smooth}}$.}
    \label{fig:ablation}
\end{figure}

We present a qualitative ablation study on how geometry constraints can influence the reconstruction quality in Fig.~\ref{fig:ablation}. The Eikonal loss $\mathcal{L}_{\textrm{eikonal}}$ and SDF loss $\mathcal{L}_{\textrm{sdf}}$ play important roles in improving geometry recovery, while the visibility loss $\mathcal{L}_{\textrm{visible}}$ and smoothness loss $\mathcal{L}_{\textrm{smooth}}$ help refine the surface.

\section{Conclusion}
This paper presents a novel neural-field-based approach, called EndoSurf, to reconstruct the deforming surgical sites from stereo endoscope videos. Our approach overcomes the geometry limitations of prior work by utilizing a neural SDF field to represent the shape, which is constrained by customized regularization techniques. In addition, we employ neural deformation and radiance fields to model surface dynamics and appearance. To disentangle the appearance learning from geometry, we incorporate normals and locations as extra clues for the radiance field. Experiments on public datasets demonstrate that our method achieves state-of-the-art results compared with existing solutions, particularly in retrieving high-fidelity shapes.

\subsubsection{Acknowledgments}

This research is funded in part via an ARC Discovery project research grant (DP220100800). 

\bibliographystyle{splncs04}
\bibliography{main}

\newpage

\title{EndoSurf: Neural Surface Reconstruction of Deformable Tissues with Stereo Endoscope Videos - Supplementary Material }

\titlerunning{EndoSurf 3D Reconstruction}

\author{\textsuperscript{*}Ruyi Zha\inst{1} \Envelope \and
\textsuperscript{*}Xuelian Cheng\inst{2,4,5} \and
Hongdong Li\inst{1} \and
Mehrtash Harandi\inst{2} \and
Zongyuan Ge \inst{2,3,4,5,6}
}
\index{Ruyi, Zha}
\index{Xuelian, Cheng} 
\index{Hongdong, Li}
\index{Mehrtash, Harandi} 
\index{Zongyuan, Ge}
\authorrunning{R. Zha et al.}
%
\institute
{Australian National University, Canberra, Australia, 
\and Faculty of Engineering, Monash University, Melbourne, Australia
\and Faculty of IT, Monash University, Melbourne, Australia 
\and AIM for Health Lab, Monash University, Melbourne, Australia 
\and Monash Medical AI, Monash University, Melbourne, Australia 
\and Airdoc-Monash Research Lab, Monash University, Melbourne, Australia 
}

\maketitle              

\section{Dataset information}
\begin{table}[H]
\centering
\tiny
\caption{Details of endoscope datasets used in the experiments.}
\label{tab:dataset}
\resizebox{\columnwidth}{!}{
\begin{tabular}{c|ccccccc}
\toprule
Scene ID & Dataset & \# of train & \# of test & Image resolution & Depth source & Scene & Camera \\ \midrule
ENDONERF-cutting & ENDONERF & 136 & 20 & $640\times512$   &  STTR-L & Dynamic  & Fixed  \\ 
ENDONERF-pulling  & ENDONERF & 55  & 8  & $640\times512$   &  STTR-L & Dynamic  & Fixed  \\
\midrule
SCARED-d1k1     & SCARED  & 86  & 13  & $1280\times1024$ &  Structured light & Static  & Moving \\
SCARED-d2k1     & SCARED   & 77  & 11  & $1280\times1024$ &  Structured light & Static  & Moving \\
SCARED-d3k1    & SCARED  & 72  & 11 & $1280\times1024$ &  Structured light & Static  & Moving \\
SCARED-d6k1     & SCARED  & 70  & 10 & $1280\times1024$ &  Structured light &Static  & Moving \\
SCARED-d7k1     & SCARED  & 71  & 10  & $1280\times1024$ &  Structured light & Static  & Moving \\ \bottomrule
\end{tabular}
}
\end{table}


\section{Hyperparameters}
\begin{table}[H]
\centering
\caption{Hyper-parameters used in this work.}
\label{tab:hyperparameters}
\resizebox{\columnwidth}{!}{%
\begin{tabular}{@{}cc|cc|cc|cc@{}}
\toprule
Parameters & Value & Parameters & Value & Parameters & Value & Parameters & Value \\ \midrule
\multicolumn{2}{c|}{Rendering} & \multicolumn{2}{c|}{Training} & \multicolumn{2}{c|}{Loss} & \multicolumn{2}{c}{Radiance field} \\ \midrule
Number of rays per iteration & 1024 & Number of iterations & 100000 & Color loss weight & 1.0 & MLP depth & 8 \\
Number of points for coarse sampling & 32 & Learning rate & 0.005 & Depth loss weight & 1.0 & MLP width & 256 \\
Number of points for fine sampling & 32 & Learning rate decay & 0.05 & SDF loss weight & 1.0 & Skip layer & 4 \\
Fine sampling steps & 4 & Warm up iteration & 5000 & Visibility loss weight & 0.1 & Encoding for location & 10 \\ \cmidrule(r){1-4}
\multicolumn{2}{c|}{Deformation field} & \multicolumn{2}{c|}{SDF field} & Eikonal loss weight & 0.1 & Encoding for direction & 4 \\ \cmidrule(r){1-4}
MLP depth & 8 & MLP depth & 8 & Surface smoothness loss & 0.1 & Activation & ReLU \\ \cmidrule(l){7-8} 
MLP width & 256 & MLP width & 256 & Surface neighbour radius & 0.1 & \multicolumn{2}{c}{Deviation network} \\ \cmidrule(l){5-8} 
Skip layer & 4 & Skip layer & 4 & \multicolumn{2}{c|}{Mesh extraction} & Initial value & 0.3 \\ \cmidrule(l){5-8} 
Encoding & 6 & Encoding & 6 & Meshgrid resolution & 128 & \multicolumn{2}{c}{SDF initialization} \\ \cmidrule(l){7-8} 
Activation & ReLU & Activation & Softplus & Marching cubes threshold & 0 & Bias & 0.8 \\ \bottomrule
\end{tabular}%
}
\end{table}

\section{Threshold validation}
\begin{figure}[H]
    \centering
    \includegraphics[width=\linewidth]{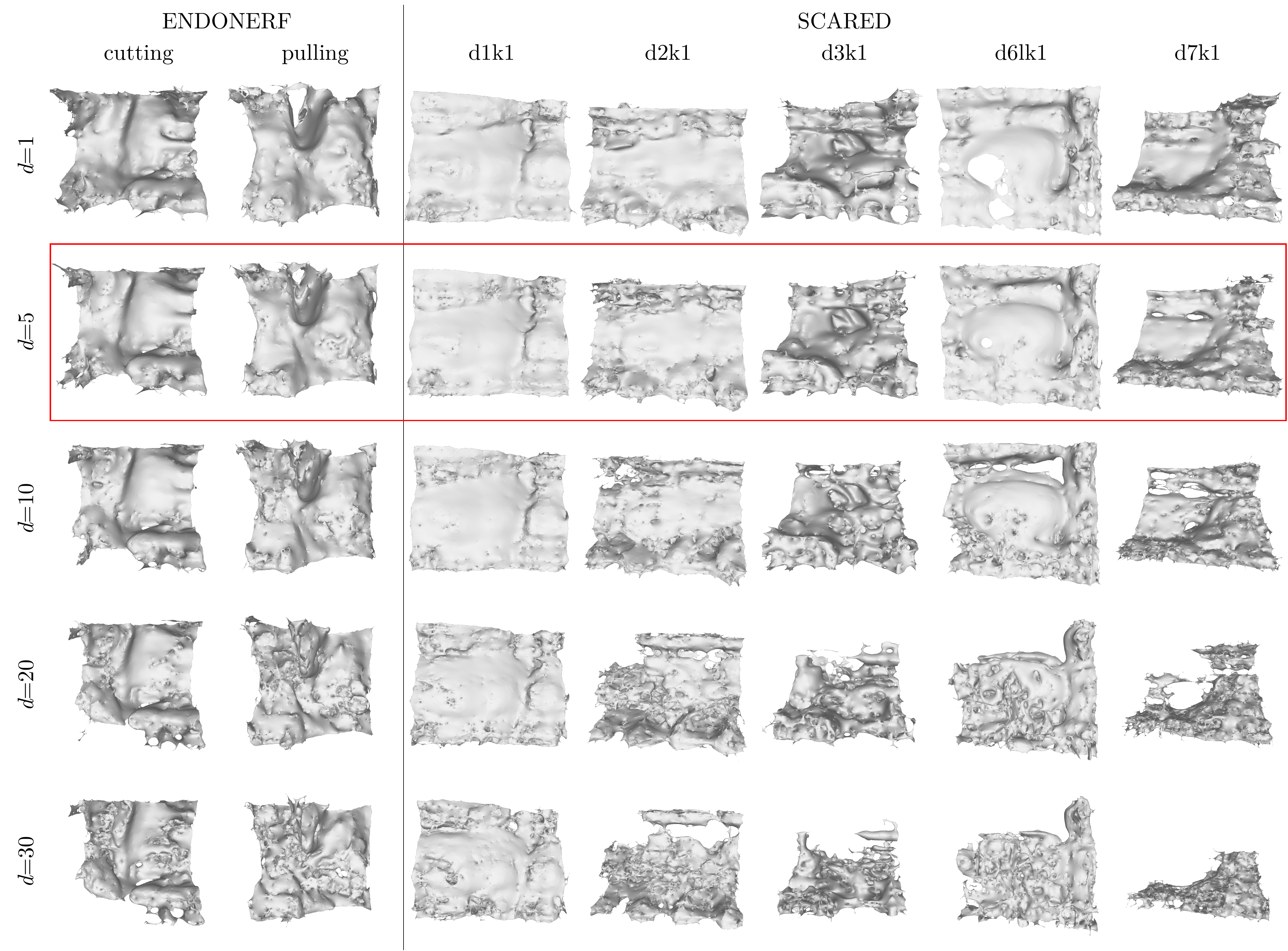}
    \caption{Validation of the threshold selection for EndoNeRF. We run marching cubes under different thresholds $d$ for all datasets and visually check if the reconstructed mesh approximates the surface of interest. Results under $d=5$ demonstrate the best reconstruction quality.}
    \label{fig:threshold}
\end{figure}

\end{document}